\documentclass[runningheads]{llncs}
\usepackage[T1]{fontenc}
\usepackage{subcaption}
\usepackage{tikz}
\usetikzlibrary{positioning}
\usetikzlibrary{calc}

\usepackage{listings}
\usepackage{caption}
\usepackage{hyperref}
\definecolor{delim}{RGB}{20,105,176}
\definecolor{numb}{RGB}{106, 109, 32}
\definecolor{string}{rgb}{0.64,0.08,0.08}
\lstdefinelanguage{json}{
    showspaces=false,
    showtabs=false,
    breaklines=true,
    postbreak=\raisebox{0ex}[0ex][0ex]{\ensuremath{\color{gray}\hookrightarrow\space}},
    breakatwhitespace=true,
    basicstyle=\ttfamily\small,
    upquote=true,
    morestring=[b]",
    stringstyle=\color{string},
    literate=
     *{0}{{{\color{numb}0}}}{1}
      {1}{{{\color{numb}1}}}{1}
      {2}{{{\color{numb}2}}}{1}
      {3}{{{\color{numb}3}}}{1}
      {4}{{{\color{numb}4}}}{1}
      {5}{{{\color{numb}5}}}{1}
      {6}{{{\color{numb}6}}}{1}
      {7}{{{\color{numb}7}}}{1}
      {8}{{{\color{numb}8}}}{1}
      {9}{{{\color{numb}9}}}{1}
      {\{}{{{\color{delim}{\{}}}}{1}
      {\}}{{{\color{delim}{\}}}}}{1}
      {[}{{{\color{delim}{[}}}}{1}
      {]}{{{\color{delim}{]}}}}{1},
}

\usepackage{graphicx}
\begin{document}
\title{Benchmarking Large Language Models on Floating-Point Error Classification}
%
%
\author{Lisa Taldir\inst{1} \and
Muhammad Ahmad Saeed\inst{1} \and
David Defour\inst{1} \and
Pablo de Oliveira Castro\inst{2}\and
Eric Petit\inst{3}}

\authorrunning{L. Taldir, M. A. Saeed, D. Defour, P. de Oliveira and E. Petit.}
%
\institute{Université de Perpignan via Domitia, Perpignan, France \\
\and Université Paris-Saclay, UVSQ, France \\ \and Intel Corp., Portland, USA}
\maketitle              
\begin{abstract}
This paper investigates the capability of Large Language Models (LLMs) to statically classify floating-point errors in software code across six categories: cancellation, comparison, division by zero, overflow, underflow and NaN. We introduce InterFLOPBench, a benchmark of 90 short C kernels (10–30
lines) with 1130 test samples designed to evaluate 17 LLMs across these error types. The evaluation framework treats floating-point error detection as a multi-label classification problem and employs the F1-score metric to measure performance. Results demonstrate that recent models achieve an overall F1-score above 0.90. Performance varies across error categories, from explicit error patterns such as division by zero (average F1-score: 0.83) to more subtle numerical phenomena such as underflow (0.65) and cancellation (0.61). While state-of-the-art LLMs show strong overall capability in static floating-point error classification, their reduced accuracy on subtle numerical phenomena highlights the need for hybrid analysis.

\end{abstract}

\section{Introduction}
\label{sec:intro}

The emergence of Large Language Models (LLMs) has generated significant interest, as they are models capable of performing complex tasks such as code generation \cite{jiang2024}, question answering \cite{yue2025}, and mathematical reasoning \cite{wang2025}. 
While this potential remains largely unexplored in many areas, such
as programming involving floating-point numbers, a few recent
studies have started to investigate this direction.
Mhatre et al. \cite{mhatre2026llmsbugscodeevaluation} evaluate LLMs on broad software bug detection, including numerical issues, while Nguyen et al. \cite{nguyen2026} leverage LLMs to generate inputs that expose floating-point inconsistencies and to rewrite expressions to improve numerical stability. In this paper, we take a complementary perspective and investigate how well LLMs can detect, explain, and classify floating-point arithmetic errors in software code. We consider 
six categories of computational anomalies: \textit{Division by Zero}, 
where a finite non-zero number is divided by zero; \textit{NaN} (Not a 
Number), resulting from invalid operations such as $0/0$ or $\sqrt{-1}$; 
\textit{Overflow}, where results exceed the maximum representable magnitude; 
\textit{Underflow}, where an operation yields a subnormal number too small 
to be represented as normalized; \textit{Cancellation}, arising when two 
nearly equal numbers are subtracted, which is particularly problematic in 
iterative algorithms; and \textit{Comparison}, where two floating-point 
numbers are tested for equality.

To support this investigation, we introduce InterFLOPBench, a multi-label 
benchmark comprising crafted C code samples that elicit specific numerical 
anomalies, paired with structured input datasets that trigger these errors. 
We use FPChecker~\cite{fpchecker2019} as the ground-truth oracle, 
complemented by Herbgrind~\cite{sanchez2017finding} and manual inspection of disagreements, leading to findings detailed in Section~\ref{sec:eval}.

Our contributions are: (1) InterFLOPBench, a new multi-label benchmark for FP error classification; (2) a systematic evaluation of 17 LLMs spanning Oct 2023 - Feb 2026; (3) a prompt engineering study showing up to +0.28 F1 improvement through precise error definitions and requesting explanations; and (4) an analysis of LLM vs. oracle disagreements revealing ground-truth limitations.

\section{Related work}
\label{sec:related_work}

\subsection{Existing Approaches to Floating-Point Error Detection}

Floating-point errors are notoriously difficult to detect due to their silent nature and strong dependence on execution context, compiler optimizations, and input data. Over the years, a wide range of tools and techniques have been proposed to assist developers in identifying such errors, broadly falling into static analysis, dynamic analysis, and hybrid approaches.
\vspace{-0.3cm}
\paragraph{Static analysis}
Static analysis tools inspect source code or intermediate representations without execution. Polyspace employs abstract interpretation to detect a variety of runtime errors—including overflow, division by zero, underflow, and NaN propagation—in languages such as C/C++ and Ada \cite{polyspace}. Tools like PRECiSA focus on static round-off error bounds and provide formal proofs \cite{precisaNASA}.
\vspace{-0.3cm}
\paragraph{Dynamic analysis and runtime monitoring}
Dynamic tools instrument programs at runtime to observe floating-point behaviour on concrete inputs. Valgrind and its extensions (e.g., Verrou~\cite{fevotte2019debugging}) provide floating-point instrumentation for error diagnosis, while Herbgrind identifies root causes of numerical issues during execution \cite{sanchez2017finding}. FPChecker detects exceptions and exceptional conditions such as NaNs, infinities, underflow, and cancellation in HPC codes \cite{fpchecker2019}.
Probabilistic and statistical methods have been proposed to assess numerical quality via stochastic perturbations. CADNA (Control of Accuracy and Debugging for Numerical Applications) implements discrete stochastic arithmetic (CESTAC) to estimate round-off error propagation and detect instabilities during execution \cite{jezequel2008cadna}. Verificarlo integrates Monte Carlo Arithmetic (MCA) into the LLVM compilation pipeline to simulate rounding error effects post-optimization, providing insights into numerical accuracy including the impact of optimizations \cite{denis2015verificarlo} and of mixed-precision computations~\cite{Chatelain2019automatic}. FLDLib is a lightweight runtime library that intercepts floating-point operations to detect anomalies such as NaNs and overflows with minimal intrusion into source code \cite{fldlib}.
\vspace{-0.3cm}
\paragraph{Benchmarks and community standards}
While individual tools report numerical anomalies, FPBench provides a standardized suite of benchmarks and accuracy measures for the floating-point research community. FPBench defines the FPCore benchmark format, associated metadata, and error metrics that allow diverse tools to be compared and composed on a common set of floating-point problems \cite{fpbench.org}. By enabling consistent evaluation across tools, FPBench supports reproducible benchmarking and fosters research into numerical analysis, optimization, and verification methods.
\vspace{-0.45cm}
\paragraph{Limitations and research gap}
Despite this rich tool ecosystem, many traditional techniques require expert interpretation, either rely on comprehensive test suites (dynamic tools) or over-approximate results with potential false positives (static tools). Moreover, they are not designed to leverage semantic understanding of code patterns that lead to floating-point anomalies. The recent advent of Large Language Models (LLMs) opens the possibility of interpreting code and detecting floating-point errors based on learned representations rather than handcrafted analysis — a new direction that this paper explores.

\subsection{Large Language Models in Code Analysis}

The emergence of LLMs as tools for code understanding represents a major advancement in automated software analysis. Trained on large-scale corpora of source code and natural language, LLMs have demonstrated strong capabilities in a wide range of programming-related tasks, including code generation, bug detection, program repair, and semantic code understanding. Their effectiveness has been demonstrated on widely adopted benchmarks such as HumanEval \cite{humaneval} and MBPP \cite{MBPP}, which primarily assess functional correctness and reasoning over short programming tasks.

The transformer-based architecture underlying modern LLMs enables the modeling of long-range dependencies within code through self-attention mechanisms \cite{vaswani2023attentionneed}. This capability is particularly well-suited to program analysis, where semantic relationships often span multiple functions, control-flow paths, or data dependencies. As a result, LLMs have been successfully applied to tasks such as vulnerability detection, API misuse identification, and logical or semantic bug detection \cite{pearce2021}. More recent work has explored the use of LLMs for program analysis tasks traditionally addressed by static or formal methods, including invariant inference, test-case generation, and type reasoning, suggesting that LLMs can serve as flexible, lightweight complements to existing analysis pipelines \cite{zhang2024survey,yang2025surveyllmbasedautomatedprogram}.

Unlike traditional static analysis techniques based on abstract interpretation or symbolic execution, which rely on sound but conservative approximations of program behavior, LLMs operate by learning implicit patterns from data rather than enforcing formally defined semantic rules. This distinction enables LLMs to generalize across coding styles and programming paradigms but also implies the absence of soundness guarantees. Consequently, LLM-based analysis is best viewed as complementary to formal methods: while abstract interpretation can prove the absence of certain classes of errors, LLMs may assist in identifying likely error patterns, prioritizing suspicious code regions, or providing early feedback during development.

However, numerical computing and floating-point arithmetic in particular, poses challenges that fundamentally differ from those addressed by conventional bug detection. Floating-point errors are governed by the IEEE-754 standard and often arise from subtle interactions between numerical representations, rounding modes, and input-dependent execution paths. Errors such as overflow, underflow, division by zero, or NaN generation are typically triggered only under specific value ranges, while others—such as cancellation or unsafe comparisons—require reasoning about relative magnitudes and numerical sensitivity rather than syntactic correctness.

Numerical analysis further involves concepts such as conditioning, numerical stability, convergence properties, and precision trade-offs, which are rarely explicit in source code and may be underrepresented in the training data of general-purpose LLMs. For instance, cancellation errors occur when subtracting nearly equal floating-point numbers, a phenomenon that cannot be detected through local pattern matching alone but requires reasoning about value distributions and algorithmic structure. Similarly, floating-point comparisons for equality may be semantically incorrect despite being syntactically valid, depending on accumulated rounding errors.

To date, only limited work has investigated the ability of LLMs to reason about floating-point semantics or IEEE-754–specific anomalies. While recent studies have begun to apply LLMs to this domain, their focus differs significantly from the approach proposed here. Recent work~\cite{wang2025llm4fp} has utilized LLM-guided strategies to generate diverse floating-point programs for testing purposes, utilizing the model to produce code that exposes inconsistencies rather than to detect bugs in existing software. Similarly, other research has integrated LLMs with human-in-the-loop refinement to verify floating-point arithmetic~\cite{mohanty2025formal}; however, this methodology operates within the context of low-level hardware Register Transfer Level (RTL) design, whereas our work targets high-level programming languages. We propose to explore whether LLMs can detect and correctly classify floating-point errors such as overflow, underflow, NaN propagation, division by zero, cancellation, and unsafe comparisons. In this study we systematically evaluate various LLMs using a dedicated benchmark and a runtime-based ground-truth oracle.

\section{Experimental Methodology}
\label{sec:eval}

\subsection{Benchmark}

InterFLOPBench, available at \url{https://github.com/interflop/InterflopBench}, covers seven floating-point error categories: cancellation, overflow, underflow, NaN, division by zero, comparison inaccuracies, and a no-error baseline, corresponding to FPChecker-identified types. Each category comprises multiple benchmark files with standardized format: a C source file with deliberately crafted code that elicits specific floating-point errors (Listing \ref{cfile}), and a corresponding JSON metadata file providing structured input datasets designed to trigger target numerical anomalies (Listing \ref{jsonfile}).

\begin{lstlisting}[caption=Kernel producing a NaN if cos(x) $\leq$ y., language=C++, numbers=none, escapechar=!, linewidth=\linewidth, label=cfile, basicstyle=\scriptsize, captionpos=b] 
double kernel(double x, double y){
    double oscillator = cos(x) - y;
    double logValue = log(oscillator);
    double result = (exp(logValue/2.0) + log(fabs(oscillator)
                + 1.0)) / (1.0 + fabs(logValue));
    
    return result;
}
\end{lstlisting}

\begin{samepage}
\begin{lstlisting}[caption=JSON file containing inputs that trigger or not a NaN error., numbers=none, escapechar=!, linewidth=\linewidth, label=jsonfile, language=json, basicstyle=\scriptsize, captionpos=b] 
{"ordered_inputs": ["x", "y"],
 "ordered_outputs": ["result"],
 "datasets": [
   {"errors": ["no_error"],!%
           \tikz[remember picture]\node (e) {};%
           !
    "inputs": [
       {"x":0.0, "y":-0.5}, {"x":1.57, "y":-0.2}
    ]},
   {"errors": ["nan"], !%
           \tikz[remember picture]\node (f) {};%
           !
    "inputs": [
       {"x":0.0, "y":1.5}, {"x":3.14, "y":1.1}]}]}
\end{lstlisting}
\noindent
\begin{tikzpicture}[remember picture, overlay,
    every edge/.append style = {->, thick, >=stealth,
                               gray, dashed, line width = 1.5pt},
    every node/.append style = {align = center, minimum height = 9pt,
                               font = \bfseries, fill = blue!20},
    text width = 1.8cm]
  \node [right = 4.3cm of e, yshift=-0.4cm] (E)
    {\small Errors predicted by \texttt{\bfseries FPChecker} for the inputs below.};
  \coordinate (Ewest) at ($(E.north west)!0.37!(E.south west)$);
  \coordinate (Esouth) at ($(E.south west)!0.12!(E.north west)$);
  
  \draw (Ewest) edge [out=180, in=0] (e.east);
  \draw (Esouth) edge [out=180, in=0] (f.east);
\end{tikzpicture}
\end{samepage}

\noindent
\begin{minipage}[t]{0.55\linewidth}
  \raggedright
  There are a total of 90 different kernels and 1\,130 samples, with 1\,307 error occurrences (errors are not mutually exclusive -- e.g., overflow often accompanies division by zero). In Table~\ref{tab:interflopbench}, ``\texttt{Occurrences}'' counts the total instances of each error category.
\end{minipage}\hfill
\begin{minipage}[t]{0.4\linewidth}
  \centering
  \small
  \begin{tabular}{lc}
    \hline
    \bfseries Category & \bfseries Occurrences \\
    \hline
    Cancellation       & 208 \\
    Comparison         & 249 \\
    Division by Zero   &  70 \\
    NaN                &  55 \\
    Overflow           & 152 \\
    Underflow          &  93 \\
    No Error           & 480 \\
    \hline
    \bfseries Total    & \bfseries 1\,307 \\
    \hline
  \end{tabular}
  \captionof{table}{InterFLOPBench error category distribution.}
  \label{tab:interflopbench}
\end{minipage}

\subsection{Experimental Setup}

\subsubsection{Models}
We evaluate 17 LLMs spanning Oct 2023–Feb 2026 on InterFLOPBench, covering diverse sizes, architectures, and reasoning capabilities (Table~\ref{tab:llm}).

\begin{table}[httb]
\centering
\small
\begin{tabular}{|p{3.2cm}|p{1.3cm}|p{1.5cm}|p{3.5cm}|p{2.1cm}|}
\hline
\textbf{Model} & \textbf{Size} & \textbf{Date} & \textbf{Reasoning} & \textbf{Architecture} \\
\hline
Mistral 7B~\cite{jiang2023mistral7b} & 7B & Oct 2023 & No & Dense \\
\hline
GPT-4o~\cite{openai2024gpt4ocard} & -- & May 2024 & CoT-capable & MoE \\
\hline
Llama 3.1~\cite{grattafiori2024llama3herdmodels} & 8B & Jul 2024 & No & Dense \\
\hline
DeepSeek R1~\cite{guo2025deepseekr1} & 8B, 32B & Jan 2025 & RL-enhanced CoT + thinking tokens & MoE \\
\hline
Gemini 2.0 Flash~\cite{comanici2025gemini25pushingfrontier} & -- & Jan 2025 & CoT-capable & Dense \\
\hline
Gemma 3~\cite{gemmateam2025gemma3technicalreport} & 4B, 12B, 27B & Mar 2025 & No & Dense \\
\hline
DeepSeek R1T~\cite{klagges2025assemblyexpertslineartimeconstruction} & 685B & Apr 2025 & RL-enhanced CoT + test-time scaling & MoE \\
\hline
Phi-4 Reasoning~\cite{abdin2025phi4reasoningtechnicalreport} & 14B & Apr 2025 & Explicit thinking tokens + CoT fine-tuning & Dense \\
\hline
Gemini 2.5 Flash~\cite{comanici2025gemini25pushingfrontier} & -- & Jun 2025 & CoT-capable & MoE \\
\hline
DeepSeek R1T2~\cite{klagges2025assemblyexpertslineartimeconstruction} & 671B & Jul 2025 & RL-enhanced CoT + inference-time scaling & MoE \\
\hline
gpt-oss~\cite{openai2025gptoss120bgptoss20bmodel} & 20B, 120B & Aug 2025 & Native CoT + configurable reasoning effort & MoE \\
\hline
Qwen 3~\cite{yang2025qwen3technicalreport} & 32B & Sep 2025 & Hybrid: Thinking/Non-thinking modes + CoT & Dense \\
\hline
Qwen 3.5~\cite{qwen2026qwen35omni} & 27B & Feb 2026 & Hybrid: Thinking/Non-thinking modes + CoT & Dense \\
\hline 
\end{tabular}
\caption{LLM models evaluated on InterFLOPBench, with sizes, release dates, reasoning mechanisms, and architecture types.}
\label{tab:llm}
\vspace{-1cm}
\end{table}
\subsubsection{Prompts}
The baseline methodology involves providing individual LLM with both C source code and JSON metadata and requesting that they classify floating-point errors for each input value. \textit{Note: The JSON metadata files are anonymized so that LLMs only see the inputs and not the errors predicted by FPChecker}. For each test case, the LLM must identify which of the six possible error categories are present, or confirm that no error occurs.

Three distinct prompts were evaluated on Qwen 3 32b, a model with configurable thinking capability (also named \emph{Chain-of-Thought} or CoT), and Gemma 3 27b, a non-reasoning model, to investigate the impact of detailed numerical analysis instructions on performance. 
\textit{Prompt 1} provides the list of the six floating-point error categories, inputs and a structured JSON response format with '\texttt{input}', '\texttt{status}' (error/no error) and '\texttt{error\_type}' fields.
\textit{Prompt 2} extends \textit{Prompt 1} with the definition of each error category added and
\textit{Prompt 3} completes \textit{Prompt 2} with '\texttt{detailed\_explanation}', requiring the model to justify each classification.

\begin{figure}
    \centering
    \begin{minipage}{0.49\linewidth}
        \centering
        \includegraphics[width=\linewidth]{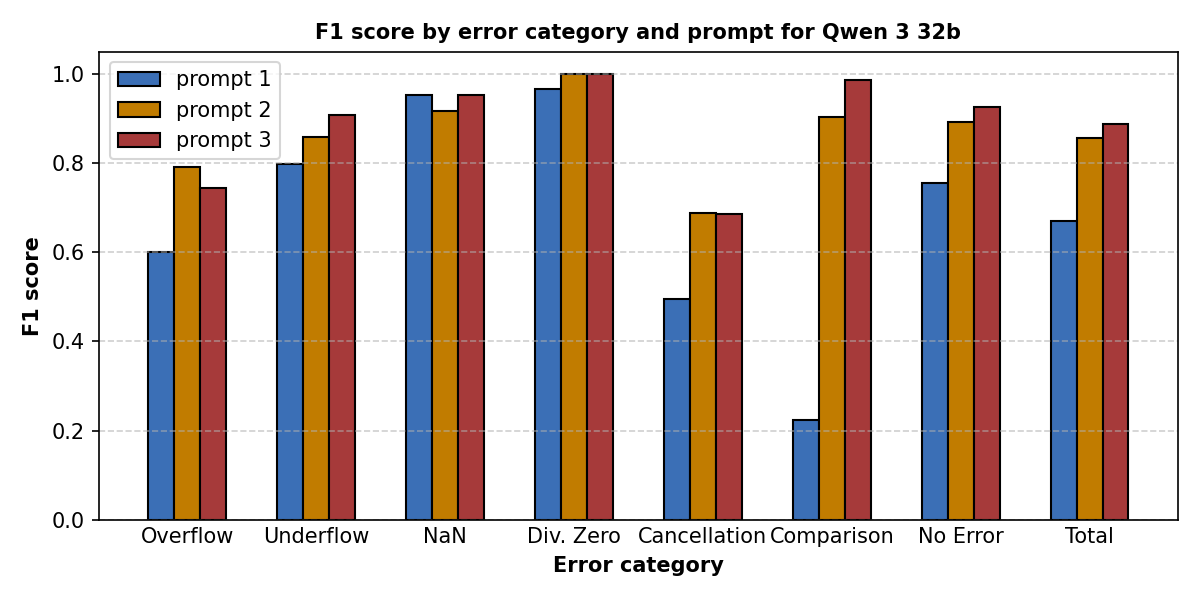}
        \caption{\small Impact of Prompt Engineering on Qwen 3 32b (F1: 0.67 $\rightarrow$ 0.86 $\rightarrow$ 0.89).}
        \label{fig:prompt_qwen}
    \end{minipage}
    \hfill
    \begin{minipage}{0.49\linewidth}
        \centering
        \includegraphics[width=\linewidth]{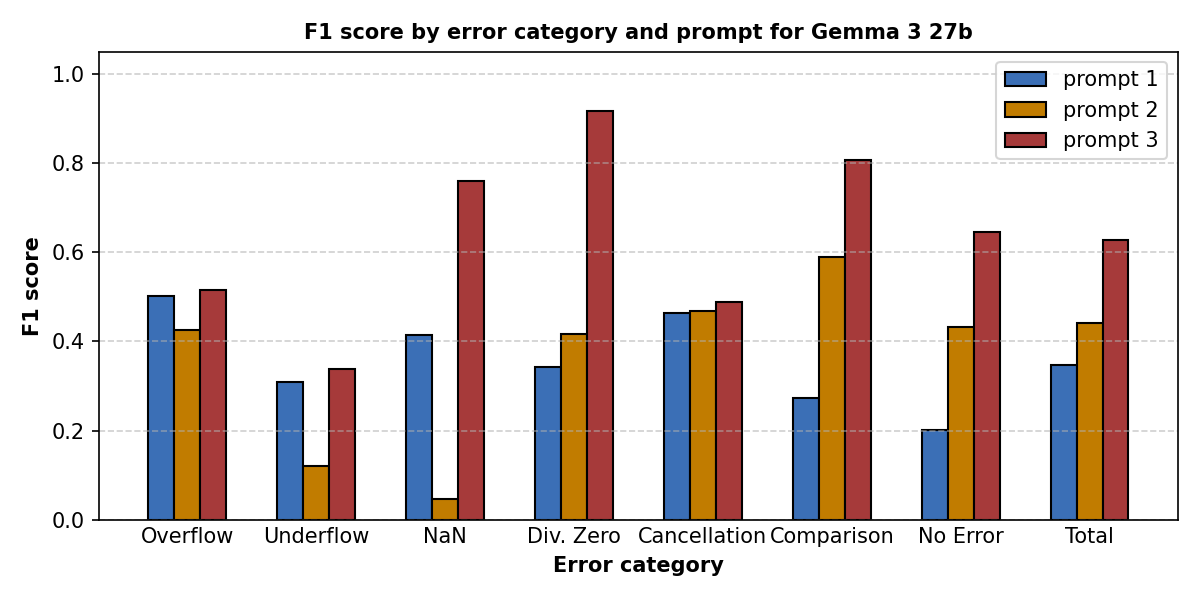}
        \caption{\small Impact of Prompt Engineering on Gemma 3 27b (F1: 0.35 $\rightarrow$ 0.44 $\rightarrow$ 0.63).}
        \label{fig:prompt_gemma3}
    \end{minipage}
    \vspace{-0.5cm}
\end{figure}

Figures \ref{fig:prompt_qwen} and \ref{fig:prompt_gemma3} reveal contrasting behaviours across the two models. For Qwen 3 32b, performance degrades when prompts lack explicit definitions for overflow, cancellation, and comparison. For Gemma 3 27b, however, adding those same definitions causes degradation: the comparison-specific rules bias the model toward over-predicting this category, compromising classification accuracy for other floating-point error types.
The addition of the \texttt{detailed\_explanation} field in \textit{Prompt 3} reconciles both cases. For Qwen 3 32b, the gain over \textit{Prompt 2} is modest (+0.03 F1), as its internal CoT already provides implicit step-by-step reasoning. For Gemma 3 27b, which lacks internal CoT, the improvement is substantial (+0.19 F1): requiring the model to explicitly justify each classification allows it to self-correct its over-prediction of comparison-related errors. These results suggest that explicitly eliciting step-by-step reasoning is particularly beneficial for models without internal CoT. \textit{Prompt 3} is therefore retained for all subsequent evaluations. Listing \ref{llm_response} shows a representative response from gpt-oss-120b.

\begin{lstlisting}[caption=Extract of the response from gpt-oss-120b on an InterFLOPBench kernel for NaN (Listing \ref{cfile}-- \ref{jsonfile})., numbers=none, escapechar=!, linewidth=\linewidth, label=llm_response, language=json, basicstyle=\scriptsize, captionpos=b] 
{"error_analysis": [
  {
    "input": {"x": "0.0", "y": "-0.5"},
    "detailed_explanation": "oscillator = cos(0.0) - (-0.5)c= 1.0 + 0.5 = 1.5 > 0, so log(oscillator) is defined and all subsequent operations produce finite numbers. No floating-point error occurs.",
    "status": "no_error",
    "error_type": ["no_error"]
  },{
    "input": {"x": 0.0, "y": 1.5},
    "detailed_explanation": "cos(0) = 1, so oscillator = 1 - 1.5 = -0.5. The argument to log() is negative, which is outside the domain of the natural logarithm. In C, log(negative) yields NaN. Consequently logValue becomes NaN, propagating through the rest of the expression and producing a NaN result.",
    "status": "error",
    "error_type": ["NaN"]}]}
\end{lstlisting}

\subsection{Ground Truth Generation and Hardening}
\label{sec:fpcheck}

To ensure a rigorous evaluation, we employ a hardened configuration of FPChecker, complemented by Herbgrind and careful manual inspection of disagreements, as the ground truth oracle. FPChecker detects the exact error type occurring at the instruction level such as NaNs, infinities, underflow, and cancellation during execution. Rather than redefining the ground truth to match Herbgrind's output-accuracy metric, we preserved FPChecker's exception-based semantics and addressed its instrumentation blind spots through targeted code hardening. Thanks to a cross‑validation with Herbgrind and manual inspection we discovered 3 FPChecker limitations which directly motivated the following modifications: 
\vspace{-0.35cm}
\paragraph{Instruction Isolation}
FPChecker cancellation checks do not handle FMA operations. Therefore, we compile all kernels with \texttt{-ffp-contract=off}. This flag disables Fused Multiply-Add (FMA) generation, ensuring that FPChecker instruments individual addition and multiplication operations separately and can detect cancellations that would otherwise be hidden inside a fused operation.
\vspace{-0.35cm}
\paragraph{Explicit Inequality Checks}
FPChecker detects strict equality comparisons (\texttt{==}) but overlooks inequality checks (\texttt{!=}) between floating-point variables, which are equally unsafe. We systematically rewrite all instances of \texttt{a~!=~b} as \texttt{!(a~==~b)}, to force the validation of the comparison logic.
\vspace{-0.35cm}
\paragraph{Library Function Propagation}
FPChecker does not instrument \texttt{math.h} library functions (e.g., domain errors in \texttt{log} or \texttt{sqrt}). To capture these errors, we append \texttt{+~0.0f} to function results, forcing value propagation through instrumented arithmetic and triggering detection.
A minor exception remains: this method is ineffective for \texttt{expf(-x)} when $x \geq 104$ since the result is a direct zero rather than a subnormal number. This affects one excluded kernel containing 10~samples.

In each case, the fix was applied to the code so that FPChecker itself detects the previously missed exception, with no change to the ground truth taxonomy.
Two residual disagreement patterns were left unresolved by design. First, benign cancellations, where FPChecker fires but Herbgrind confirms no output damage, are retained as error-labelled entries, consistent with FPChecker's exception-based definition. Second, silent precision loss, where FPChecker reports no\_error but Herbgrind detects 26–62 bits of degradation, was not promoted to an error label, to avoid conflating exception occurrence with output accuracy. Notably, on both disagreement patterns, LLMs tend to side with Herbgrind's implicit assessment, sensing a numerical risk even outside the strict exception range; this causes a slight underestimation of LLM capability in the reported F1-scores, as discussed in Section~\ref{sec:oracle_refinement}.

\subsection{Evaluation Metrics}

Floating-point error detection is treated as a multi-label classification problem since each input can potentially exhibit multiple error types simultaneously. The label space comprises seven categories: cancellation, underflow, overflow, NaN, division by zero, comparison, and no error. As illustrated in Table~\ref{tab:classification}, each prediction is evaluated per label in terms of true positive (TP), false positive (FP) and false negative (FN). The F1-score is then used as the primary metric, and reports both per-category and overall (micro-averaged) scores.
\[
    F1 = 2 \times \frac{TP}{2TP + FP + FN}
\]
\vspace{-0.35cm}
\begin{table}[h!]
    \centering
    \footnotesize
    \setlength{\tabcolsep}{3pt}  
    \renewcommand{\arraystretch}{1.1}
    \begin{tabular}{|p{2cm}|p{2.5cm}|p{2.5cm}|p{4.2cm}|}
        \hline
        \textbf{Scenario} & \textbf{True} & \textbf{Predicted} & \textbf{Impact} \\
        \hline
        Correct & overflow & overflow & \textcolor{green!50!black}{TP} overflow \\
        \hline
        Missing & overflow & no error & \textcolor{red!60!black}{FN} overflow + \textcolor{red!60!black}{FP} no error \\
        \hline
        False alarm & no error& overflow & \textcolor{red!60!black}{FP} overflow + \textcolor{red!60!black}{FN} no error\\
        \hline
        Under-pred & \{overflow, NaN\} & \{overflow\} & \textcolor{green!50!black}{TP} overflow + \textcolor{red!60!black}{FN} NaN \\
        \hline
        Over-pred & \{NaN\} & \{overflow, NaN\} & \textcolor{green!50!black}{TP} NaN + \textcolor{red!60!black}{FP} overflow \\
        \hline
    \end{tabular}
    \caption{TP, FP, and FN outcomes in the multi-label classification setting.}
    \label{tab:classification}
    \vspace{-0.7cm}
\end{table}

\section{Results}
\label{sec:results}

\subsection{InterFLOPBench Results}
\begin{figure*}[h!]
    \centering    
    \makebox[\linewidth][c]{\includegraphics[width=1.3\linewidth]{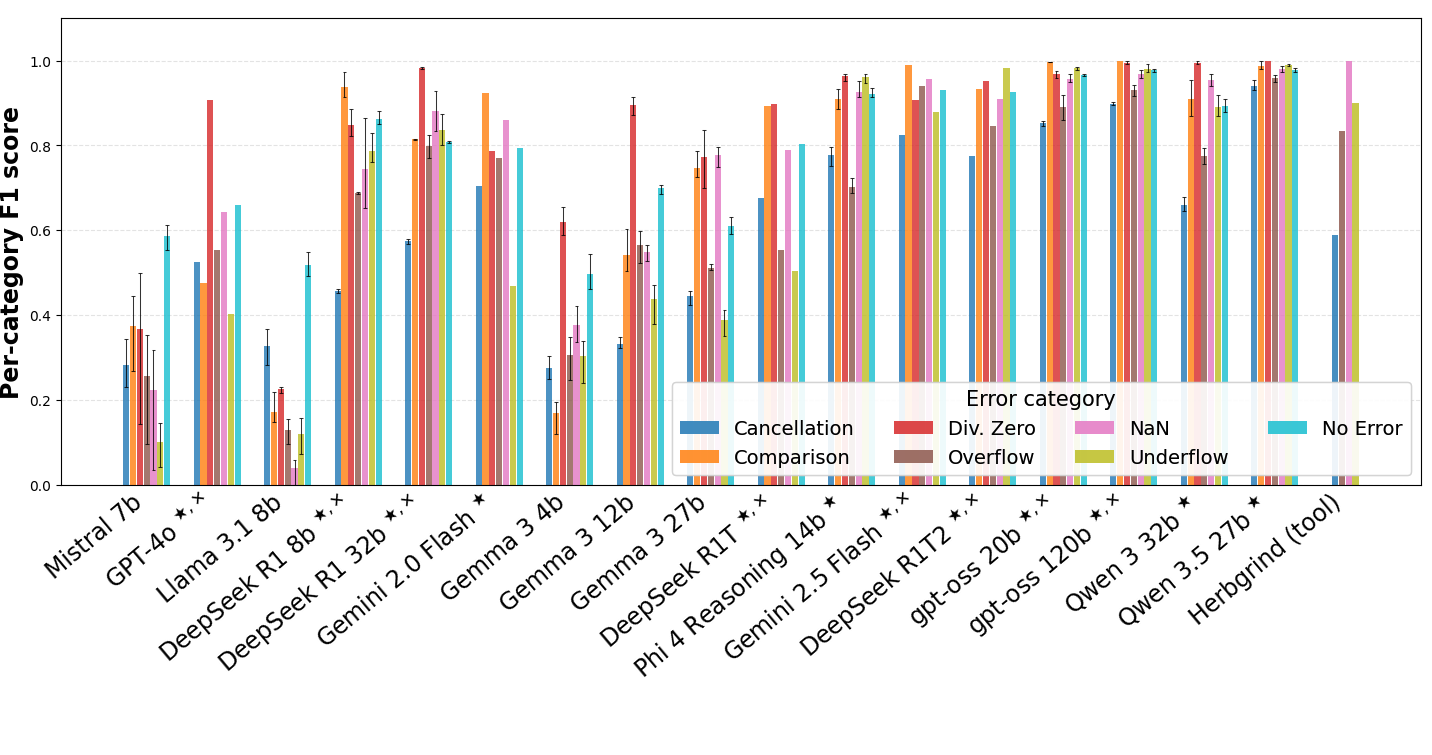}}
        \vspace{-0.75cm}
        \caption{\small Per-category F1-score across 17 LLMs sorted by release date from the oldest (left), to the latest (right). \ensuremath{\star}~Chain-of-Thought (CoT) reasoning; \ensuremath{\times}~Mixture-of-Experts (MoE) architecture. Full per-benchmark disagreement analysis is available online {\small \href{https://html-preview.github.io/?url=https://github.com/interflop/InterflopBench/blob/version_0.1/ground_truth_analysis_table.html}{https://github.com/interflop/InterflopBench/ground\_truth\_analysis\_table.html}}.} 
    \label{fig:per_category_f1_score}
\end{figure*}

Fig. \ref{fig:per_category_f1_score} presents the overall F1-scores per category for the 17 evaluated models. It shows a chronological improvement in floating-point error detection, achieving over 90\% F1-score for all types of floating-point exceptions, and with lower variability, for the recents models gpt-oss 120b, and Qwen 3.5 27b. 
Herbgrind detection rates ($\geq$10-bit output error threshold) are shown as a reference group (at the right of the figure) for categories amenable to output-accuracy validation (cancellation, overflow, NaN, underflow). 
Categories relying on structural pattern detection (Comparison, No Error) are not applicable to Herbgrind.
Low Herbgrind rates for cancellation indicate that most flagged subtractions do not propagate to output damage; high rates for Overflow/NaN reflect cases where exception-based and output-accuracy tools agree. 

\vspace{-0.35cm}

\subsection{Case Study: LULESH}

To evaluate the ability of LLMs to analyse real-world HPC applications, we introduced a guaranteed NaN ($\frac{a-a}{a-a}$) inside LULESH (see Listing \ref{lulesh}) replicating an experiment proposed in the FPChecker tutorial\footnote{\url{https://fpanalysistools.org/pearc19/slides/Module-FPChecker.pdf}}.
The complete file (60k tokens including the prompt) was provided to Gemini 2.0 Flash, Gemini 2.5 Flash and Gemini 3.0 Flash, whose context window (1M tokens) is sufficient to process the full application. We used the best performing \emph{Prompt 3}, that includes the formal definition for each floating-point error category and requests a \texttt{detailed\_explanation} field. 

\begin{lstlisting}[caption=Extract of lulesh.cu file with NaN injection., language=C++, numbers=none, escapechar=!, linewidth=\linewidth, label=lulesh, basicstyle=\scriptsize, captionpos=b] 
void CalcAccelerationForNodes_kernel(
    int numNode, Real_t *xdd, Real_t *ydd, Real_t *zdd, Real_t 
    *fx, Real_t *fy, Real_t *fz, Real_t *nodalMass){
  int tid=blockDim.x*blockIdx.x+threadIdx.x;
  if (tid < numNode){
      Real_t one_over_nMass = Real_t(1.)/nodalMass[tid];

      // injected bug
      one_over_nMass = (one_over_nMass - one_over_nMass)
                /(one_over_nMass - one_over_nMass);
      
      xdd[tid]=fx[tid]*one_over_nMass;
      ydd[tid]=fy[tid]*one_over_nMass;
      zdd[tid]=fz[tid]*one_over_nMass;
  }
}
\end{lstlisting}

All tested models correctly identified the injected NaN. Indeed, the expression $\frac{a-a}{a-a}$ is a $0/0$ pattern that can be detected syntactically, without requiring computations or multi-step reasoning.

To further evaluate reasoning robustness, we tested the models under two progressively more complex NaN injected bugs.
\textit{Test 2} Insertion of $(a+b)/(c+d)$ with fixed constants such that $a=-b$ and $c=-d$.
\textit{Test 3} Same expression as Test 1, but where $a,b,c,d$ are computed variables (still satisfying $a=-b$ and $c=-d$).
Table \ref{tab:gemini_lulesh} summarizes the results. Gemini 2.5 and 3.0 correctly detect the injected NaN in all scenarios. However, the older Gemini 2.0 Flash fails the most complex case (Test 3). While the model correctly infers that the denominator is zero, it fails to recognize that the numerator is also zero, resulting in a misclassification as "Division by Zero." This highlights the limitations of earlier model generations in performing static analysis on chains of operations with computed dependencies.

\begin{table}[h]
\centering
\begin{tabular}{|p{3cm}|p{2.5cm}|p{1.5cm}|p{3cm}|}
\hline
\textbf{Model} 
& \textbf{Direct $\frac{a-a}{a-a}$} 
& \textbf{Test 2} 
& \textbf{Test 3} \\
\hline
Gemini 2.0 Flash & NaN & NaN & Division by Zero \\
Gemini 2.5 Flash & NaN & NaN & NaN \\
Gemini 3.0 Flash & NaN & NaN & NaN \\
\hline
\end{tabular}
\caption{NaN detection robustness in LULESH under progressively complex formulations.}
\label{tab:gemini_lulesh}
\end{table}

\section{Discussions}
\label{sec:discussions}

\subsection{Why LLMs Succeed at Detecting Floating-Point Errors}

LLMs are effective at static detection of numerical anomalies because they learn semantic regularities from massive corpora of real code (common numerical idioms, typical bad floating-point patterns, and known pitfalls such as catastrophic cancellation or unsafe comparisons). 
This learned pattern recognition is crucial for detecting anomalies like cancellation, comparison of floats, or NaN-producing patterns that are implicit in the code structure rather than syntactically obvious.

Beyond pattern matching, LLMs can reason about quantities: performing simple computations on floating-point operations via chain-of-thought, inferring whether a value accumulates through a loop, or detecting mismatches between mathematical intent and the implemented floating-point computation. 
Many numerical bugs are not illegal operations but numerically unstable formulations; LLMs can infer the intended mathematical expression and flag the discrepancy.

Static analysis tools typically require concrete value ranges to diagnose an issue; when bounds are unknown, inputs come from external sources, or behavior depends on runtime conditions, they either remain silent or conservatively over-approximate.
LLMs, by contrast, can still reason probabilistically about data magnitudes and flag potentially unsafe operations: for instance, identifying that \texttt{exp(x)} will overflow for large positive~$x$, or that a loop multiplying by a constant greater than one will eventually exceed \texttt{DBL\_MAX}, even without knowing the exact input range.
This ability to assign qualitative risk rather than binary verdicts makes them effective front-end detectors, surfacing suspicious code regions for further analysis by other tools.

A distinctive strength is \emph{interpretability}: unlike runtime tools that report an error type and location, LLMs can classify the anomaly, explain the underlying numerical mechanism, suggest safer reformulations, and reference known numerical analysis principles.
This makes them valuable as front-end detectors and educational tools, even when they are not the final arbiters of correctness.

\subsection{Where LLMs Fail}

We noticed that performance degrades on subtle error categories. Cancellation and underflow, i.e. where the error depends on relative magnitudes or accumulated rounding across iterations, remain the hardest to detect (average F1: 0.62 and 0.65 respectively, versus 0.83 for division by zero). 
Intermediate overflow is another blind spot: when the final result is representable but the computation path overflows (e.g., squaring \texttt{expm1(x)} for large~$x$), LLMs tend to reason about endpoints rather than intermediate values.

Model size is a strong predictor for these harder categories. 
Larger and more recent models track longer dependency chains and reason about multi-step floating-point effects; 
smaller models often catch division by zero and NaN but miss subtle cancellation or overflow-through-iteration.
Beyond scale, training methodology matters: models that underwent explicit reasoning reinforcement (e.g., GRPO or chain-of-thought reinforcement learning)  tend to outperform models of comparable or even larger size, as illustrated by Phi 4 Reasoning (14b) surpassing GPT-4o in our results. 
The ability to carry out multi-step numerical reasoning, rather than data exposure alone, is the key differentiator.

\subsection{Disagreement Analysis and Oracle Refinement}
\label{sec:oracle_refinement}

The hardening protocols described in Section~\ref{sec:fpcheck} were not designed a priori: they were discovered through a systematic analysis of discrepancies between one of the best-performing model (gpt-oss-120b) and the standard FPChecker configuration.

Out of 1\,130 test samples, gpt-oss-120b disagreed with the oracle on 86~instances. Manual inspection and Herbgrind (to consolidate NaN detection for example) revealed that in 28 of these cases, the LLM was correct and surprisingly the oracle remains silent. On these cases, the LLM correctly identified: implicit cancellation risks in FMA operations, unsafe inequality comparisons (\texttt{!=}), and domain errors within \texttt{math.h} library calls and the oracle did not.

Each of these findings led directly to one of the three hardening protocols applied to the benchmark (Section~\ref{sec:fpcheck}). This demonstrates a significant result: LLMs appear to be able to detect semantic floating-point errors that evade standard runtime instrumentation, effectively serving as auditors or complementary tools for conventional dynamic analysis tools. The F1~scores reported in this paper are therefore measured against the hardened ground truth, as with a standard FPChecker configuration, many of the LLM's apparent false positives would in fact be true positives that the dynamic tool did not instrument.

A complementary Herbgrind analysis reveals the opposite failure mode: cases where FPChecker reports no\_error yet the output has lost 26–62 bits due to silent rounding accumulation (log(1+x), Rump's example\footnote{Available at: \url{https://github.com/interflop/InterflopBench/blob/version_0.1/examples/03_rump/bench.c}}). These samples were deliberately not relabelled, as promoting them would conflate exception occurrence with output accuracy and make the evaluation target ambiguous. As a consequence, the F1 scores reported here slightly underestimate LLM capability on this specific class: models that correctly flag these patterns are scored as false positives, even though they outperform FPChecker on arguably the most dangerous errors.

\subsection{Costs}

Token consumption varies significantly across model families and directly affects inference cost. Non-reasoning models such as Gemma 3 generate 500–1 000 tokens per sample, limited to the requested JSON output. Reasoning models such as DeepSeek R1 and Qwen 3.5 produce an additional internal chain-of-thought, raising token counts to 5 000–10 000 per sample -- a 5–10× overhead. Given that InterFLOPBench comprises 1 130 test samples, this translates to a non-negligible cost difference at scale. The performance gains reported in Section \ref{sec:results} suggest this overhead is justified for the hardest error categories (cancellation, underflow), where reasoning models outperform their non-reasoning counterparts most markedly; for simpler categories such as division by zero or NaN, non-reasoning models offer a more cost-effective alternative.

\subsection{Scope and Future Directions}

The current evaluation has a bounded scope: C-only code, single-prompt single-turn interactions, and isolated kernel functions. 
Extending InterFLOPBench to Fortran, Python with NumPy, or mixed-language HPC applications would broaden applicability. 
Evaluating multi-turn or agentic pipelines (where an LLM iteratively refines its analysis or decomposes a large codebase into smaller units) may yield higher detection rates on real-world complex software systems.
These limitations highlight concrete directions for progress, both in improving the scope of runtime instrumentation tools and in developing LLM-based approaches—such as agentic pipelines where LLMs can leverage existing FP debugging tools, 
complement their coverage, and eventually surpass standalone approaches.

\section{Conclusion}
\label{sec:conclusion}

This paper presents a systematic evaluation of 17~LLMs on floating-point error detection and classification using InterFLOPBench, a new benchmark comprising 90~kernels and 1\,130 test samples in 6 error categories. The best-performing models 
achieve scores greater than 0.9, demonstrating that modern LLMs can reliably detect a wide range of floating-point anomalies from source code alone. 

Prompt engineering plays a significant role: providing precise error definitions improves F1 by up to~0.19 for reasoning models, while explicit CoT yield useful gains for interpretability (additional +0.03 for reasoning models, and +0.19 for non-reasoning models). Performance varies substantially across error categories—explicit errors, such as division by zero and NaN, are well handled, while subtle phenomena, like cancellation and underflow, remain challenging.

Detailed analysis of disagreements between the best model and the ground-truth tool FPChecker reveals cases where the LLM was correct and the oracle silent, notably for FMA-related cancellation and comparisons involving inequality operators. These findings suggest that LLMs are not merely matching runtime tools but can complement them by covering blind spots in dynamic instrumentation.
LLMs are not yet meant to replace formal verification or dynamic analysis, but they are already effective as a semantic filter to locate suspicious code, act as a classification engines to categorize errors, and explanation engines to detail why an operation is numerically unsafe. Combined with their ability to suggest more stable reformulations, they offer a promising complement to existing floating-point debugging workflows.

%
%
%
\bibliographystyle{splncs04}
\bibliography{biblio}

\end{document}